\documentclass[letterpaper, 10 pt, conference]{ieeeconf}

\usepackage{algorithm}
\usepackage[noend]{algpseudocode}
\usepackage{amsmath,amssymb}
\usepackage{cite}
\usepackage{graphicx}
\usepackage{subcaption}
\usepackage[hang,flushmargin]{footmisc}

\IEEEoverridecommandlockouts % This command is only needed if you want to use the \thanks command

\title{\LARGE \bf
Information-Collection in Robotic Process Monitoring:\\
An Active Perception Approach}

\author{Martin A. Sehr$^{1,\ast}$, Wei Xi Xia$^{1,\ast}$, Prithvi Akella$^{2}$, Juan Aparicio Ojea$^{1}$, Eugen Solowjow$^{1}$% <-this % stops a space
\thanks{$^1$ Siemens, Berkeley, CA 94704, USA.}
\thanks{$^2$ Department of Mechanical and Civil Engineering, California Institute of Technology, Pasadena, CA 91125, USA.}
\thanks{$\ast$ The first two authors contributed equally to this work.}}

\begin{document}
\maketitle
\thispagestyle{empty}
\pagestyle{empty}

%%%%%%%%%%%%%%%%%%%%%%%%%%%%%%%%%%%%%%%%%%%%%%%%%%%%%%%%%%%%%%%%%%%%%%%%%%%%%%%%
\begin{abstract}
Active perception systems maximizing information gain to support both monitoring and decision making have seen considerable application in recent work.
In this paper, we propose and demonstrate a method of acquiring and extrapolating information in an active sensory system through use of a Bayesian Filter.
Our approach is motivated by manufacturing processes, where automated visual tracking of system states may aid in fault diagnosis, certification of parts and safety; in extreme cases, our approach may enable novel manufacturing processes relying on monitoring solutions beyond passive perception.
We demonstrate how using a Bayesian Filter in active perception scenarios permits reasoning about future actions based on measured as well as unmeasured but propagated state elements, thereby increasing substantially the quality of information available to decision making algorithms used in control of overarching processes.
We demonstrate use of our active perception system in physical experiments, where we use a time-varying Kalman Filter to resolve uncertainty for a representative system capturing in additive manufacturing.
\end{abstract}

%%%%%%%%%%%%%%%%%%%%%%%%%%%%%%%%%%%%%%%%%%%%%%%%%%%%%%%%%%%%%%%%%%%%%%%%%%%%%%%%
\section{Introduction}
In modern manufacturing, a wide array of thermomechanical processes are used in production and modification of parts and objects of various materials.
The purposes of these processes are diverse, including hardening of structures, increase in corrosion resistance, additive manufacturing and many others.
Regardless of their widespread application in industry, complex thermomechanical manufacturing processes remain difficult to control precisely for a variety of reasons, many of which relate to insufficient knowledge of process states for either re- and pro-active decision making, often going in hand with little to no use of adequate process models.
A common challenge with thermomechanical processes is that even slight changes in temperature above certain thresholds can result in severe damage or even loss of parts.
Hence, accurate real-time estimation of temperature fields in parts is of great interest.

Unfortunately, only few applications have access to the measurement data needed to accomplish this.
In some cases, one could attempt to distribute stationary sensors over a workspace and compose their measurements to reconstruct desired information.
While possible in principle, this method is often impractical for a variety of reasons, including workspace constraints such as occlusions and moving parts, unit costs per sensor added or excessive exposure to heat, vibrations and other process factors.
One can circumvent these issues by dynamically repositioning a mobile sensor and gathering all data required from that single sensor.
Naturally, this alternative allows incorporating additional, stationary or mobile sensors installed in the workspace.
The nature of this problem typifies those encountered in active sensing, where re-orientation of a single sensor such as a camera taking images from various angles provides the same if not richer sensory signals as compared to multiple stationary sensors.
%An active sensor can also help to capture time-variant processes by always providing data with high information content. 

\begin{figure}[tb]
\centering
\includegraphics[width = \columnwidth]{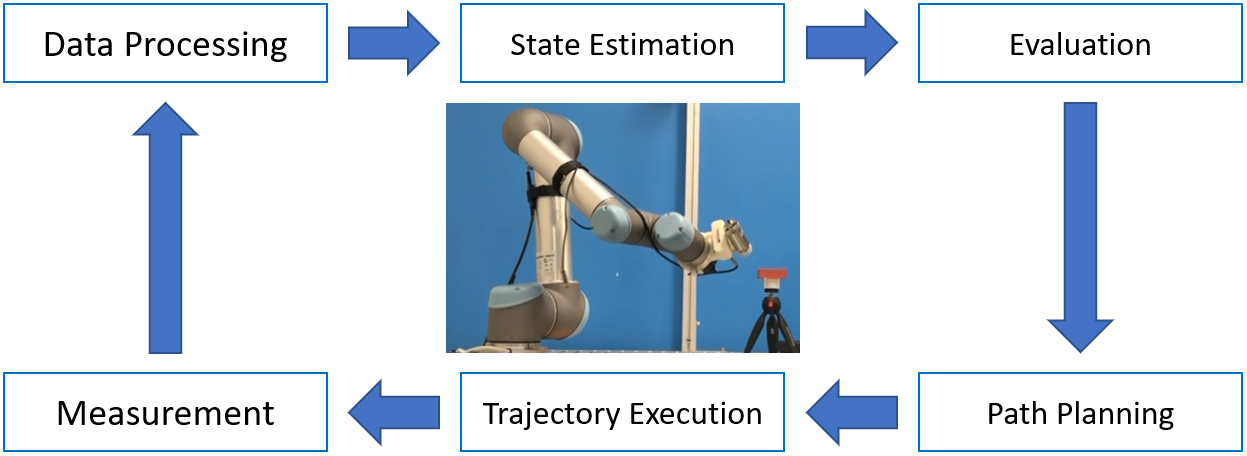}
\caption{Active sensing for distributed physical phenomena, repeating indefinitely: (1) estimation of current state values using physical process model and Bayesian Filter, (2) evaluation of state and uncertainty information, (3) robot path planning for new sensor position capturing critical/uncertain part locations, (4) trajectory execution on robot arm, (5) measurement data capture, (6) data processing including projection and mapping of image data to process model.}
\label{fig:workflow}
\end{figure}

In order to move or re-orient the sensor so as to provide that richer signal, decision criteria are required.
A reasonable choice in practice is often to maximize information gain within the feasible active sensing field, as motivated by information-theoretic control~\cite{grocholsky2002information}.
Aim of this paper is to provide a general framework for an active-sensing system that can extrapolate from sensed information to estimate current and predict future states, and use that information to inform future sensory actions and ultimately regulate underlying physical processes.
Our algorithmic approach, visualized in Figure~\ref{fig:workflow}, is designed for active sensing of distributed physical phenomena such as temperature fields with cameras mounted on robot arms. In Section~\ref{sec:experiments}, we demonstrate our approach for a challenging monitoring problem in additive manufacturing, which requires continuous repositioning of an IR camera to capture interior and exterior temperature distributions.

\subsection{Prior work}
As detailed in~\cite{Review_Manipulation}, interest in active perception and information theoretic control have grown rapidly in recent years.
There are different ways to derive positions for information maximization. 
Optimal sensing actions are derived in~\cite{Surface-uncertainty-gaussian, ansari2016sequential, atanasov2014information} by sensing at locations with process uncertainty. 
Similarly, Gaussian process regression is used in~\cite{Surface-gaussian,kreuzer2018, hollinger2013active, duecker2019learning} to measure at locations that minimize the uncertainty of a surface model.

For more active, real-time scenarios, Bayesian Optimization techniques can help choosing sensing location via maximizing the probability of providing useful information~\cite{Bayesian-Optimisation}.
Regardless of how one chooses the information criteria, it is well accepted that maintaining a belief representation~\cite{thrun2005probabilistic} - whether by culmination of sensory information into an evolving state vector, or by updating one or multiple probability models in a non-Markovian sense - will yield richer subsequent actions.
For example, recent work in~\cite{Unscented-Localization} indicates that use of a Memory Unscented Kalman Filter is a more efficient means of Tactile Localization.
%On the other hand Graeme et al. \cite{Monte-Carlo} search a Distributed Monte Carlo Tree, which, coupled with periodic updates, helps a system of autonomous actors function more efficiently .
Combining both the Bayesian and Gaussian modelling, Extended Kalman Filters are used in~\cite{KF-trajectory} to generate optimal trajectories and find considerable success over random methods. A special case of active perception, which is also related to our contribution, is concerned with determining the next pose for a visual sensor given its previous measurements. This is often called the "next best view" problem~\cite{bircher2016receding, banta2000next, pito1999solution, potthast2014probabilistic}.

%MS CTD
\subsection{Contributions}
Following the steps outlined in~\cite{KF-trajectory}, we develop a generalized framework for the active sensing problem wherein the information used for decision-making stems from a Bayesian Filter.
Given a process-specific state-evolution matrix and state vector as well as noise statistics, we merge observations and physical process models in order to obtain high-dimensional state estimates including associated measures of uncertainty.
The main contribution of this paper is an active system that is tailored towards reducing the uncertainty about dynamically varying scalar fields on objects in manufacturing processes.
The system can be used to automatically and actively gather concurrent state information from thermomechanical processes at runtime.
This application illustrates the feasibility of joint active sensing and closed-loop state estimation of scalar fields on physical objects.

\section{Methodology}
This paper discusses a particular class of state estimation problems for nonlinear autonomous
%\footnote{The absence of control inputs does not affect the state estimation process and is without loss of generality in the context of this paper.} 
 systems of the form
\begin{align}\label{eqn:state}
    x_{k+1} &= f_k(x_k,w_k), &
    x_0 &\in\mathbb{R}^{n},
\end{align}
where $x_k$ denotes the state vector, $w_k$ the process noise vector and $f_k$ the state evolution function for $k\in\mathbb{N}_0$.
The class of estimation problems tackled in this paper is motivated by successively repositioning sensor equipment to resolve the state of system~\eqref{eqn:state}, leading to output equation
\begin{align}\label{eqn:output}
    y_{k} &= g_k(x_k,v_k),
\end{align}
where $y_k$ denotes the system output vector, $v_k$ the measurement noise vector, and $g_k$ the output function for $k\in\mathbb{N}_0$.
The particular output function $g_k$ at time instance $k$ depends on the current location of the sensing equipment, which is selected based on the statistics of the Bayesian Filter~\cite{simon2006optimal} used to estimate the state vector,
\begin{align}\label{eqn:BF1}
    \pi_k &= \frac{\operatorname{pdf}(y_k\mid x_k)\pi_{k\mid k-1}}{\int\operatorname{pdf}(y_k\mid x_k)\pi_{k\mid k-1}d x_k}, \\ \label{eqn:BF2}
    \pi_{k+1\mid k} &= \int\operatorname{pdf}(x_{k+1}\mid x_k)\pi_k d x_k,
\end{align}
where $\pi_k$ denotes the probability density function (pdf) of $x_k$ given past measurement data and initial prior density, $\pi_{0\mid -1}$.
This leads to the following abstract routine:
\begin{algorithm}[h]
    \caption{Active Perception System} \label{alg:activeperception}
    \begin{algorithmic}[1]
    \For{$k \in \mathbb{N}_0$}
    \State Based on $\pi_{k\mid k-1}$, position sensors
    \State Based on sensor positions, derive $g_{k}$
    \State Measure system output $y_k$ via~\eqref{eqn:output}
    \State Compute probability density $\pi_k$ via~\eqref{eqn:BF1}
    \State Compute $\pi_{k+1\mid k}$ via~\eqref{eqn:BF2}
    \EndFor
    \end{algorithmic}
\end{algorithm}

The remainder of this paper discusses the implementation of a linear variant of Algorithm~\ref{alg:activeperception}. 
However, notice that while the Bayesian Filter~\eqref{eqn:BF1}-\eqref{eqn:BF2} can be reduced to a classic Kalman Filter in the linear case, the problem structure remains as in the general, nonlinear setup.
We chose to proceed with the linear case to simplify description of our experiments in Section~\ref{sec:experiments}, which demonstrate the use of Algorithm~\ref{alg:activeperception} for a particular problem variant in which temperatures of an object are observed with an IR camera, which is relocated dynamically using a robotic to resolve uncertainty about estimated temperatures on the object surface as well as in its interior.
In the linear case, the system equation~\eqref{eqn:state} reduces to
\begin{align}\label{eqn:linstate}
    x_{k+1} &= A_k x_k + w_k,
\end{align}
the output equation~\eqref{eqn:output} to
\begin{align}\label{eqn:linoutput}
    y_k &= C_k x_k + v_k,
\end{align}
and the Bayesian Filter to a classic Kalman Filter with
\begin{align}\label{eqn:KF}
\begin{split}
P_{k|k-1} &= A_k P_{k-1|k-1}A_k^T + W_k, \\
K_k &= P_{k\mid k-1} C_k^T (V_k + C_k P_{k\mid k-1} C_k^T)^{-1}, \\
P_{k|k} &= (I - K_k C_k)P_{k|k-1}(I - K_k C_k)^T + K_k V_k K_k^T,\\
\hat{x}_{k\mid k-1} &= A_k\hat{x}_{k-1\mid k-1}, \\
\hat{x}_{k\mid k} &= \hat{x}_{k\mid k-1} + K_k(y_k - C_k\hat{x}_{k\mid k-1}),
\end{split}
\end{align}
where $\hat{x}_{k\mid k}$ denotes the state estimate vector at time $k$ using all measurements up to time $k$, $W_k$ and $V_k$ denote the respective noise covariance matrices of $w_k$ and $v_k$, $P_{k\mid k}$ the posterior state estimate covariance matrix and $K_k$ the Kalman Filter gain. 

\section{Implementation}
\label{sec:implementation}
We continue with a particular implementation of Algorithm~\ref{alg:activeperception}, aimed to estimate states of physical objects by use of a single line-of-sight sensor that can be positioned and oriented dynamically as required to resolve state estimate uncertainty.
While our implementation is based on the linear formulation captured by~\eqref{eqn:linstate}-\eqref{eqn:KF}, it can be extended seamlessly to its more general nonlinear counterpart~\eqref{eqn:state}-\eqref{eqn:BF2}. 
Similarly, given that our fundamental approach is based on a Kalman Filter, additional fixed and mobile sensors may be added without fundamental changes to our system.
In the following, we use \textsc{libigl}~\cite{libigl}, an open-source \textsc{C++} visualization library, to recreate physical objects in virtual space and derive the observation matrices $C_k$ based on sensor positions at time instance $k$, corresponding to Line~3 in Algorithm~\ref{alg:activeperception}. 

\subsection{Initialization}
\begin{figure}[tb]
\centering{
    \includegraphics[width = \columnwidth]{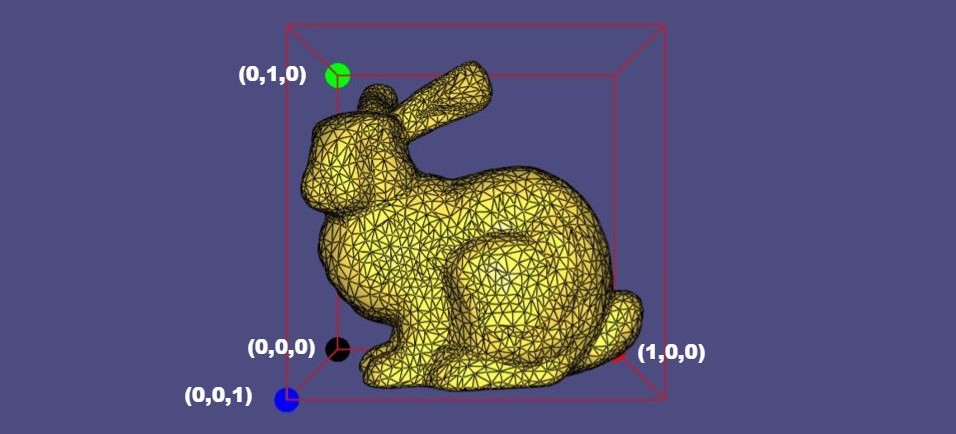}
}
\caption{
Bounding box for the Stanford Bunny~\cite{Bunny}; points annotated with Cartesian coordinates in $(x,y,z)$-notation.}
\label{fig:bounding-box}
\end{figure}

Using \textsc{libigl}, we create matrices to generate vertices and edges for given objects.
To avoid excessive state dimensions in our active perception algorithm, we select only a reduced number of object vertices to form the spatial discretization of our system state to be estimated. 
This reduction, essential especially for complex object shapes with large numbers of vertices, is based on projecting uniform points across bounding boxes of given objects (see e.g. Figure~\ref{fig:bounding-box}) onto their surface areas.
This projection is performed by selecting the vertices closest to each of the points distributed uniformly across the bounding boxes, resulting in an evenly spaced, reduced number of vertices, which form the state vector of our system model~\eqref{eqn:state} and are referred to as \textit{control points} below.\footnote{Notice that, while objects with occlusions may result in a low number of control points within said occlusions, these occlusions would typically also fall outside the field of vision of line-of-sight sensors such as cameras.
Moreover, additional control points on the surface and in the interior of a given object may be added manually.}
That is, each physical quantity (e.g. temperature) captured in our state vector is associated with a location specified by a unique control point.
Based on this system state, we generate the state evolution matrices, $A_k$, for instance via discretization of a partial differential equation, such as the heat equation in our experiments below, capturing the system dynamics, taking into account specific control point locations.

\subsection{Object Rotation and Projection}
At each measurement step of our active perception process, Lines~2-4 in Algorithm~\ref{alg:activeperception}, we move the sensor to its desired location and project a plane onto the object surfaces in line of sight.
To determine the face of the object the sensor is currently observing, we assume that the sensor is positioned at the coordinate origin and oriented pointing along positive $x$-direction; that is, the positive $x$-axis is facing from the sensor towards the center of its field of view.
Based on this fixed setup, we calculate rotation and translation matrices to re-orient the camera from its fixed rest location to its current state and perform the inverse operations on the vertices of the object observed.
Doing so orients the virtual object in the sensor reference frame.
We next average the $x$-coordinates of a slice of vertices centered on the $x$-$z$ plane, denoting their average by $\bar{x}$.
The system then classifies as observed by the sensor any control point whose $x$-coordinate lies between the average and the coordinate origin.\footnote{Notice that while this process of determining control points in line of sight assumes relatively smooth objects shapes with limited surface curvature, our implementation performed well in experiments with various objects, such as e.g. the Stanford Bunny~\cite{Bunny}.
%If needed for high-curvature objects, this face detection routine can be extended by taking into account explicitly the surface mesh edges of given objects.
}

Finally, for all observed points, the system populates a row in the current observation matrix, $C_k$, as defined in equation~\ref{eqn:linoutput}.
Algorithm~\ref{alg:Observation-algorithm} summarizes schematically the procedure to construct these observation matrices, where $\Gamma$ and $\Gamma_{\operatorname{rot}}$ denote the matrices of object vertices in default frame and sensor frame, respectively.
Furthermore, ${\mathcal{G}}$ denotes the set of $y$-coordinates explored around $y=0$, where $\bar{g}\in(0,0.5]$, 
${e}_i$ are Lagrange basis vectors in $x,y,z$-directions, respectively, $R$ corresponds to the rotation matrix required to re-orient the global $x$-axis to the positive sensor-orientation.

\begin{algorithm}[h]
    \caption{Observation Algorithm} \label{alg:Observation-algorithm}
    \begin{algorithmic}[1]
    \Procedure{Frame Change}{}
    \State $ \mathrm{Find } \, R \, | \, {x}_o = R{e}_1 $
    \State $ \Gamma_{\operatorname{rot}} = R^T(\Gamma - {x}_s)$
    \EndProcedure
    \Procedure{Partitioning}{}
    \State $y_{min} = \, \mathrm{argmin}(\{{x} \cdot {e}_2 \, | \, {x} \in \Gamma\})$
    \State $y_{max} = \, \mathrm{argmax}(\{{x} \cdot {e}_2 \, | \, {x} \in \Gamma\})$
    \State $\mathcal{G} = \{g \, | -\bar{g}(y_{max} - y_{min}) \leq g \leq  \bar{g}(y_{max} - y_{min})\}$
    \State $\bar{x} \, = \, \mathrm{avg}( \{{x} \cdot {e}_1 \, | {x} \cdot {e}_2 \in {\mathcal{G}} \, \forall \, {x} \in \Gamma  \})$
    \State $ \mathcal{P} = \{ {x} | {x} \cdot {e}_1 - \bar{x} < 0, \, \forall \, {x} \in \Gamma \} $
    \EndProcedure
    \Procedure{Observation Matrix}{}
    \For{${x} \in \mathcal{P}$}
    \State $C_k = [C_k; 0 \ldots 1 \ldots 0]$
    \EndFor
    \EndProcedure
    \end{algorithmic}
\end{algorithm}

\begin{figure*}[t]
\centering{
\begin{subfigure}[t]{\columnwidth}
\centering\includegraphics[width = \columnwidth]{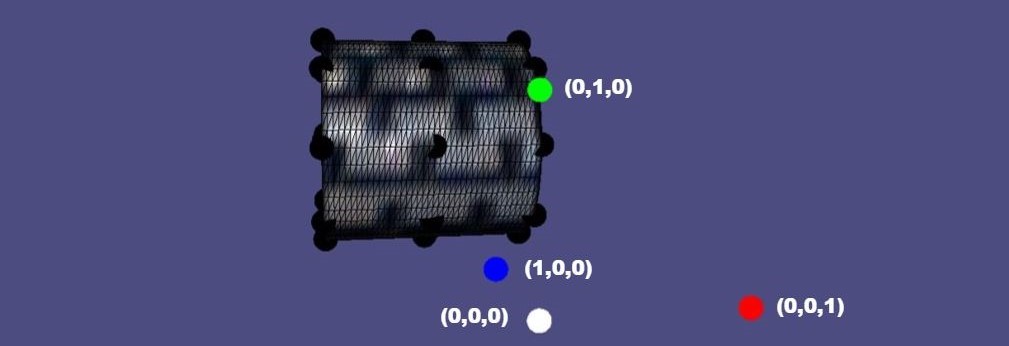}
\caption{To rotate the object into the sensor frame, we record translation and rotation matrices from fixed rest location to current sensor position (white) and orientation (positive $x$-direction), respectively.}
\end{subfigure}\hfill
\begin{subfigure}[t]{\columnwidth}
\centering\includegraphics[width=\columnwidth]{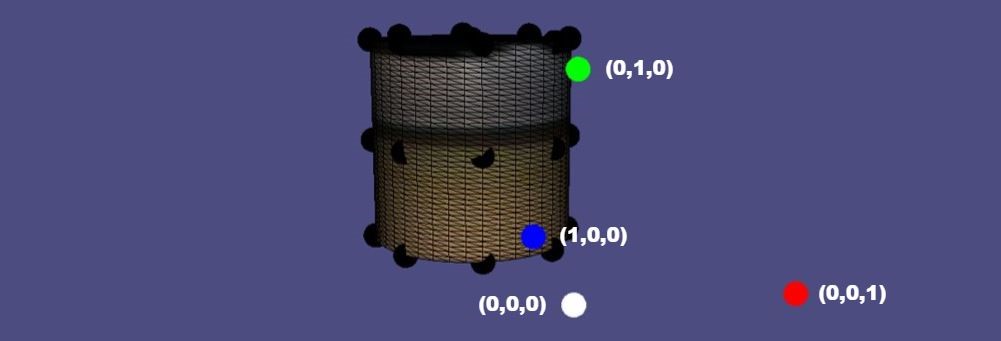}
\caption{We distinguish between observed and occluded faces of the object by calculating a local average and only projecting onto those points that are on the same side of that average as is the sensor.}
\end{subfigure}
}
\caption{Reference frame rotation and projection of sensor information onto a cylindrical object; 
control points on cylinder highlighted by black dots;
points annotated with Cartesian coordinates in $(x,y,z)$-notation.}
\label{fig:rot-proj}
\end{figure*}

After identifying the observed control points via Algorithm~\ref{alg:Observation-algorithm}, we record the data of the closest sensor information (e.g. pixels in case of a camera) as measurement value, generating an observation vector with one entry per observed control point of the object.
Based on this measurement vector, $y_k$, current state estimate vector and covariance matrix are all updated through a Kalman Filter of form~\eqref{eqn:KF}.
Figure~\ref{fig:rot-proj} highlights an example of our rotation-projection routine used to map sensor information onto the surface of a given object.

\subsection{Active Perception System}
Based on the filter estimating concurrently the state of the system, two different methods were used to prescribe new sensor positions: either the system navigated the sensor to observe the control point with the highest value (e.g. temperatures in our experiments in Section~\ref{sec:experiments}), or it navigated to the control point at which the state estimate was most uncertain.
Notice that these two variants may be augmented and combined rather arbitrarily without structural modifications to the framework presented in this study; they represent example setups for detection of extreme values and smoothing overall system uncertainty, respectively.

Regardless of this choice, new sensor positions are generated by first identifying the vector connecting the geometric center of the object, denoted $x_{\operatorname{center}}$, to the control point in question and scaling it by $\alpha > 1$ to define the displacement vector, denoted ${d}$. 
Based on that, the sensor position is defined as the sum of the displacement vector and the object's geometric center.
The sensor orientation vector, ${x}_o$, is defined as the anti-parallel unit-vector to the unit-vector in the direction of the displacement vector.
This procedure is summarized by Algorithm~\ref{alg:filter-update}, where $\mathcal{P}$ denotes the set of observed control points as per Algorithm~\ref{alg:Observation-algorithm}, $\mathcal{C}$ is the matrix of Cartesian positions of each data point in the sensor frame, $\mathcal{X}$ are the Cartesian coordinates of all object control points, $\mathcal{Y}(i,j,k)$ denotes the sensor data at location $(i,j)$ and time instance $k$, and $x_s$ and $x_c$ denote sensor position and orientation, respectively.
Positions are assigned to each sensor data point depending on the average distance of the sensor to the object, the current state estimate vector, $\hat{x}_k$, and sensor-specific width and height of observed frames.

\begin{algorithm}[htb]
    \caption{Measurement and Updating} \label{alg:filter-update}
    \begin{algorithmic}[1]
    \Procedure{Measurement and Filtering}{}
    \For{${x} \in \mathcal{P}$}
    \State ${y}_k(x_s) = \mathcal{Y}({i}, {j},k) \,\, | \,\, ({i},{j}) \, \operatorname{min} \, ||\mathcal{C}({i},{j}) - x_s||_2 $
    \EndFor
    \State ${y}_k$, $C_k \rightarrow \text{Kalman Filter} \rightarrow \hat{{x}}_k, {P}_{k|k}$
    \EndProcedure
    \Procedure{Max Value Update}{}
    \State $\mathcal{Y}_{max} \, , \, i = \operatorname{argmax}_i (\hat{{x}}_k)$
    \State ${d} = \alpha(\mathcal{X}(i,:) - x_{\operatorname{center}}) $
    \State ${x}_s = {d} +  x_{\operatorname{center}} $
    \State ${x}_o = \frac{-{d}}{||{d}||_2}$
    \EndProcedure
    \Procedure{Max Uncertainty Update}{}
    \State $U_{max} \, , \, i = \operatorname{argmax}_i (\operatorname{diag}({P}_{k|k})_i)$
    \State ${d} = \alpha(\mathcal{X}(i,:) - x_{\operatorname{center}}) $
    \State ${x}_s = {d} + x_{\operatorname{center}} $
    \State ${x}_o = \frac{-{d}}{||{d}||_2}$
    \EndProcedure
    \end{algorithmic}
\end{algorithm}
%\FloatBarrier

\section{Hardware Experiments}
\label{sec:experiments}
In order to validate our method for active perception in thermomechanical manufacturing processes, we next describe in detail a specific use case scenario and experiments for an emulated thermomechanical process.
In specific, we use our system to capture the thermal field over a part being produced by means of robotic manufacturing.
We consider a two-robot scenario where \emph{Robot A} produces a part via robotic additive manufacturing and \emph{Robot B} executes our algorithmic pipeline for active perception.

\begin{figure}[bt]
    \centering
    \includegraphics[width = \columnwidth]{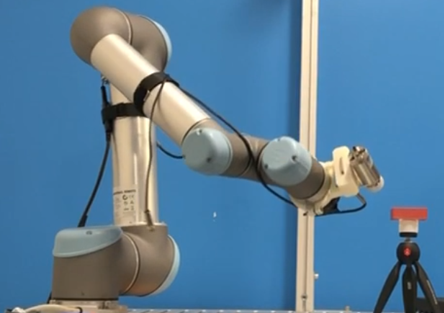}
    \caption{Experiment configuration: UR5 robot arm with mounted IR camera running active perception pipeline.}
    \label{fig:setup}
\end{figure}

For practical reasons, we emulate this scenario in a hardware-in-the-loop configuration, where the physical process guided by Robot A is simulated in virtual space while Robot B executes the motion prescribed by our algorithm in physical space with a scaled physical replica model of the part to be produced.
The physical Robot B is fully aware of the virtual world in which Robot A lives and the manufacturing process takes place, but acts in the real-world.
This setup is possible, because the manufacturing process does not depend on  Robot B's actions.
The physical components of this setup are displayed in Figure~\ref{fig:setup} with a UR5 robot arm representing Robot B and an Optris IR camera mounted facing in parallel direction with the outward pointing axis of the tool center point.
In our experiments performed for this paper, the images captured by the IR camera are discarded in favor of noisy process values governed by our detailed thermomechanical process model, which serves as ground truth for our experiments. 
However, while we discard the actual IR image data, we process the physical camera orientation via the exact projection pipeline described through Section~\ref{sec:implementation} above to extract the surface temperature data in the camera field of view from the underlying detailed thermomechanical simulation model. 

Notice that the detailed thermomechanical model used to generate ground truth data in our hardware-in-the-loop experiments is \emph{not} identical in structure to the one used by our active perception algorithm, which operates using a lower-dimensional linear time-varying model capturing the continuous deposition of material during the additive manufacturing process of interest.
Instead, we use dedicated simulation tools for thermomechanical processes to generate a detailed high-fidelity model of the underlying additive manufacturing process.
%\footnote{We use Siemens' Simcenter StarCCM+, a high-fidelity thermomechanical model using specialized simulation software for thermomechanical processes, although any similar software tool could be used instead.}
The data generated by this software is saved as a look-up table which is parsed by our hardware-in-the-loop configuration at runtime.

While this hardware-in-the-loop configuration may appear to reduce the complexity of our case study, notice that it has several key advantages over using an entirely physical manifestation of our use case scenario.
For instance, the hardware-in-the-loop configuration allows us to capture the thermomechanical process with an additional, user-specified and more fine-grained dynamic model than that processed by our Kalman Filter at runtime.
In addition, our experimental setup permits access to quantities such as interior part temperatures over time, which would be inaccessible in a complete physical version of this scenario.
This in turn allows us to analyze in detail the performance of our algorithms in the emulated additive manufacturing scenario.

\begin{figure*}
\begin{subfigure}{\textwidth}
    \centering
    \includegraphics[width = \textwidth]{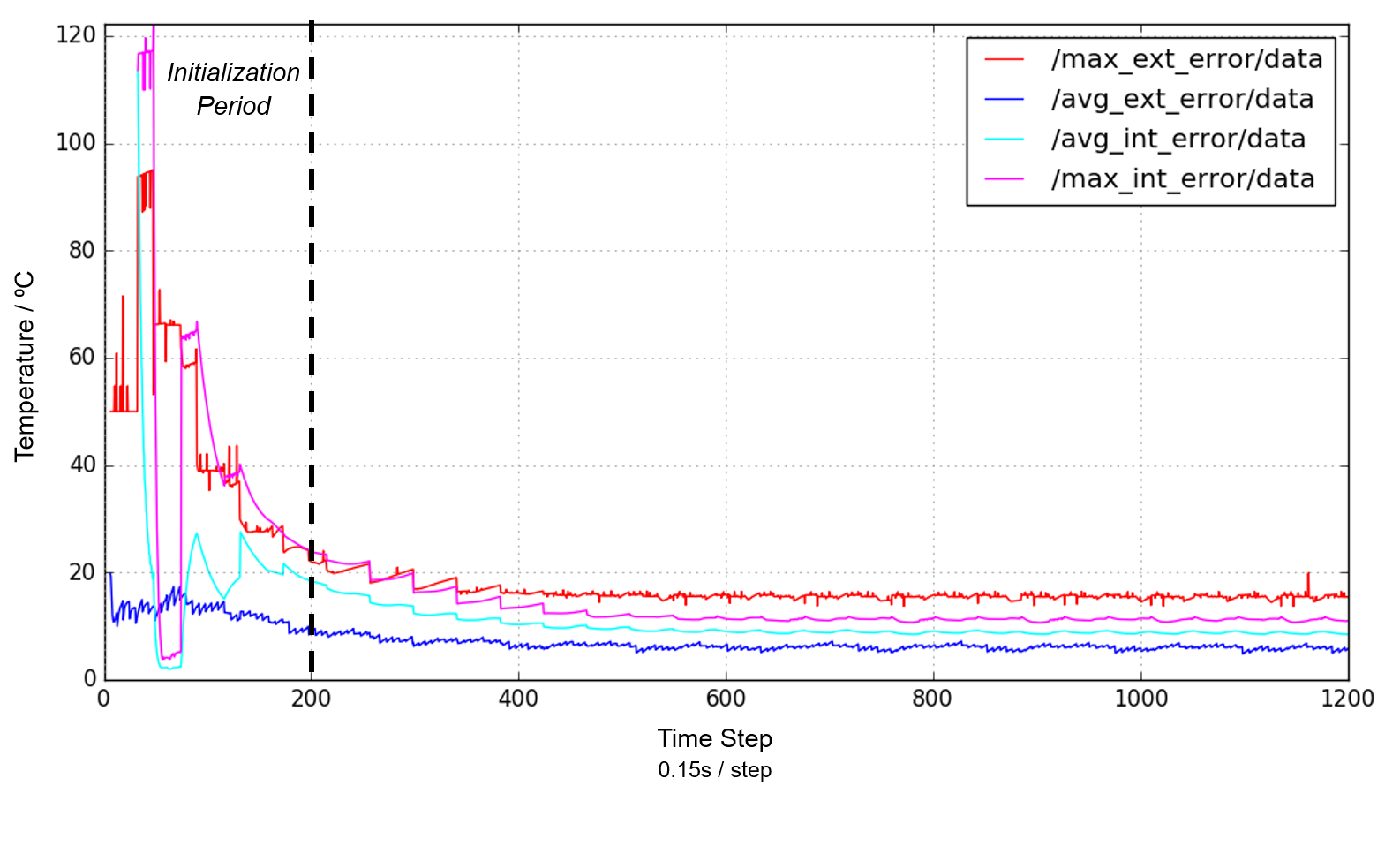}
\end{subfigure}
\begin{subfigure}{\textwidth}
    \centering
    \includegraphics[width = \textwidth]{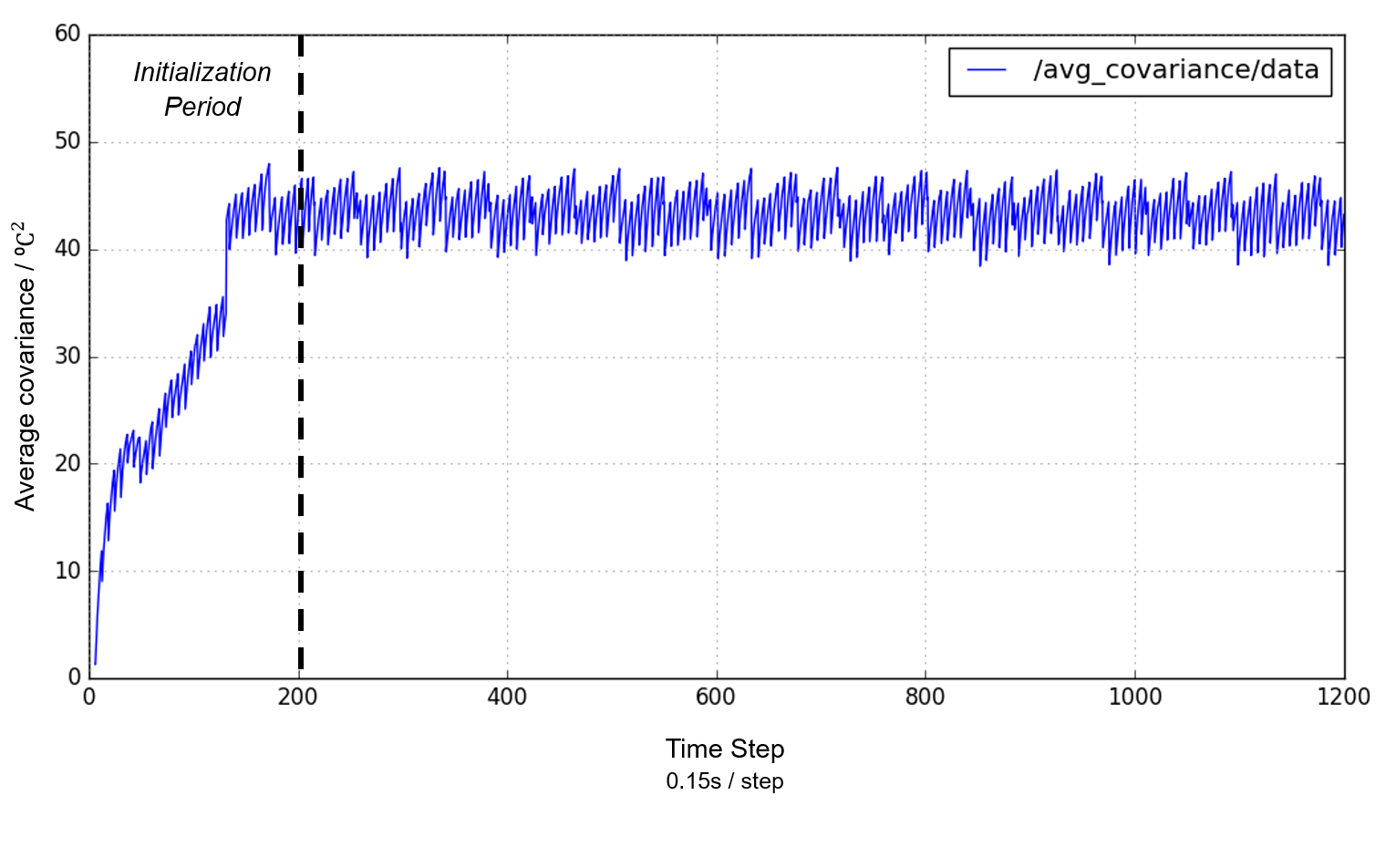}
\end{subfigure}
    \caption{Experiment data: maximum and average temperature errors at exterior (ext) and interior (int) part locations (top); average covariance across active locations. After \emph{initialization period} where initial layers are produced, temperature error and covariance curves follow steady behavior. Residual errors due to discrepancies between thermomechanical ground truth model and model processed by algorithm. Similar behavior can be expected when observing actual physical processes.}
    \label{fig:data-experiment}
\end{figure*}

The part produced in our experiments is of rectangular geometry with a concentric rectangular pocket at its center, and is produced via a tool path depositing concentric three beads per layer of material.
Notice that, while we chose a rather simple geometry for illustration, our numerical model has to capture both material deposition and heat transfer at the same time, which requires a time-varying version of our Kalman Filter~\eqref{eqn:KF}.
For numerical efficiency, our active perception algorithm captures the temperature distribution only for the layers of material produced most recently; this approximation allows us to increase spatial resolution at locations where temperatures are expected to be the highest without sacrificing computational speed.
The resulting experiment data for $4$ \emph{active} layers of material and fixed model update time step of $0.15s$ is captured by Figure~\ref{fig:data-experiment}.
Given the time required to reposition the IR camera using the robot arm, we obtain measurement data approximately every $6s$ at optimal locations in terms of temperature estimate covariance data.
After an initialization period, during which the first $4$ layers of material are deposited, the algorithm results in steady progression of both temperature errors and covariances, as expected using our algorithmic approach.
The residual errors are explained by the different modeling approaches employed to generate the ground truth data and the LTV model used to estimate the temperature field at runtime. 
While one could adjust the models to reduce these errors, we believe they are representative of what one may experience in a fully physical experiment.

\section{Conclusions}
We demonstrated the feasibility of an active sensory system for estimating scalar fields from camera frames, where only partial information is available to the sensory system.
This was shown based on a Bayesian filter to provide decision-making criteria, which depends on states that are not directly measured or cannot be measured.
For the specific use case in our experiments, we demonstrated that the active sensory system improves iteratively information gain in a challenging additive manufacturing scenario.
Given the generality of our approach, it may be used similarly in other, more complex thermomechanical scenarios and ultimately for direct process control. 

% \section*{Acknowledgements}
% This work was funded by the Advanced Robotics for Manufacturing (ARM) Institute and Siemens Corporation.

%%%%%%%%%%%%%%%%%%%%%%%%%%%%%%%%%%%%%%%%%%%%%%%%%%%%%%%%%%%%%%%%%%%%%%%%%%%%%%%%

\bibliographystyle{IEEEtran}
\bibliography{IEEEexample}

\end{document}